\documentclass{easychair}

\usepackage{doc}
\usepackage{makeidx}
\usepackage{listings}
\usepackage{appendix}
\usepackage{pdfpages}
\usepackage{graphicx}
\usepackage{amsmath, amssymb}
\usepackage{url}
\usepackage{amsmath}
\usepackage{fancyvrb}
\usepackage{textcomp}
\usepackage{authblk}
\usepackage{verbatim}
\usepackage{fancyvrb}

\begin{document}

%
\title{On the Behavioral Formalization of the Cognitive Middleware AWDRAT}


\titlerunning{Formalization of AWDRAT}

%
\author{
Muhammad Taimoor Khan\inst{1}
\and
    Dimitrios Serpanos\inst{1}
\and
   Howard Shrobe\inst{2}\\
}

\institute{
  QCRI, Qatar\\
  \email{\{mtkhan, dserpanos\}@qf.org.qa}
\and
   CSAIL, MIT, U.S.A.\\
   \email{hes@csail.mit.edu}\\
 }

\authorrunning{Khan, Serpanos and Shrobe}

\clearpage

\maketitle


%
%

\pagestyle{empty}

We present our ongoing work and initial results towards the (behavioral) correctness analysis of the cognitive middleware AWDRAT~\cite{Shrobe:2006}. Since, 
the (provable) behavioral correctness of a software system is a fundamental pre-requisite of the system's security. Therefore, the goal of the work is
to first formalize the behavioral semantics of the middleware as a pre-requisite for our proof of the behavioral correctness. However, in this
paper, we focus only on the core and critical component of the middleware, i.e. Execution Monitor which is a part of the module ``Architectural Differencer'' of AWDRAT. The role of the execution monitor is to identify inconsistencies between runtime \emph{observations} of the target system and \emph{predictions} of the specification System Architectural Model of the system. As a starting point we have defined the formal (denotational) semantics of the \emph{observations} (runtime events) and \emph{predictions} (executable specifications as of System Architectural Model); then based on the aforementioned formal semantices, we have formalized the behavior of the ``Execution Monitor'' of the middleware. The material of the parts of this paper is based on~\cite{MTK:2014}.
\enlargethispage*{1cm}

AWDRAT is a general purpose middleware system that provides survivability to any kind of new and legacy software system. In detail, the middleware checks for consistency between the target system's actual (runtime) behavior and the expected (system specification) behavior of the system, if there is the one then the diagnostic engine identifies an attack (illegal behavioral pattern) and the corresponding set of resources which were compromised during the attack. After identifying an attack, AWDRAT attempts
to repair respectively regenerate the compromised system into a safer state, if possible. The task of regeneration is based on the dependency-directed reasoning~\cite{Shrobe:1979} engine of the system that contributes to the self-organization and self-awareness of the
system by recording execution steps intrinsically states of the system and their corresponding justification (reason). Based on the Execution Monitor and the reasoning engine of AWDRAT not only the detection of known attacks is possible but also detection (resp. recovery from) the unknown attacks is also possible.


\textbf{A Specification Language of AWDRAT}: 
A specification language ``System Architectural Model'' of AWDRAT supports to specify the target system behavior based on a fairly high-level description written in a language of ``Plan Calculus''~\cite{Shrobe:1979} which is a decomposition of pre- and post- and invariant conditions for each computing component (module) of the target system. The description can be considered as an executable specification of the system. The specification is a hierarchical nesting of system's components such that input and output ports of each component are connected by data and control flow links respective specifications. Furthermore, each component is specified with corresponding pre- and post-conditions. However, the specification also includes a variety of event specifications.

In detail, the specification (System Architectural Model) of target system is described at the following two logical levels:
\begin{enumerate}
\item \emph{control level} describes the control structure of each of the component (e.g. sub-components, control flow and data flow links) which is
\begin{itemize}
\item defined by the syntactic domain ``StrModSeq'' while the control flow can further be elaborated with syntactic domain``SplModSeq''
\end{itemize}
\item \emph{behavior level} describes the actual method's behavioral specification of each of the component which is defined by the syntactic domain ``BehModSeq''.
\end{enumerate}
Furthermore, the registration of the \emph{observations} is given by the syntactic domain ``RegModSeq'' at the top of the above domains. All (four) of the aforementioned domains are the top-level syntactic domains of the System Architectural Model. For syntactic details, please see~\cite{MTK:2014}.

Based on the core idea of Lamport~\cite{Lamport:1994}, we have defined the semantics of the specification as a state relationship to achieve the desired insight of the program's behavior by relating pre- and post-states~\cite{MTK12c}. Semantically, the System Architectural Model \emph{SAM} holds in a given environment $e$ resulting in an environment $e'$ by transforming a pre-state $s$ into post-state $s'$ as defined below.

\begin{tabbing}
\textlbrackdbl \=SAM\textrbrackdbl (e)(e', s, s') $\Leftrightarrow$ 
\\\> $\forall$ \=$e_1, e_2, e_3 \in$ Environment, $s_1, s_2, s_3 \in $ State: \\\>\> \textlbrackdbl RegModSeq\textrbrackdbl (e)(e$_1$, s, inState$_\bot$(s$_1$)) $\wedge$ \textlbrackdbl StrModSeq\textrbrackdbl (e$_1$)(e$_2$, s$_1$, inState$_\bot$(s$_2$)) $\wedge$
\\\>\>\textlbrackdbl BehModSeq\textrbrackdbl (e$_2$) (e$_3$, s$_2$, inState$_\bot$(s$_3$))
$\wedge$ \textlbrackdbl SpltModSeq\textrbrackdbl (e$_3$)(e', s$_3$, s')
\end{tabbing}
For further details on the semantics, please see~\cite{MTK:2014}.

\textbf{An Execution Monitor of AWDRAT}:  
In principle, an execution monitor interprets the event stream (traces of the execution of the target system aka \emph{observations}) against the system specification (the execution of the specification is also called \emph{predictions}) by detecting inconsistencies between \emph{observations} and the
\emph{predictions}, if there is any. 
\enlargethispage*{4cm}

When the target system starts execution, an initial ``startup'' event is generated and
dispatched to the top level component (module) of the system which transforms the execution state of the component
into ``running'' mode. The component instantiates its subnetwork (of components, if there is one) and also propagates
the data along its data links by enabling the corresponding control links (if involved). When the data arrives on the input port of the component, the execution monitor checks if it is complete; if so, the execution monitor checks the preconditions of the component for the data and if they succeed, it transform the state of the component into ``ready'' mode. In case, any of the preconditions fails, it enables diagnosis engine.

After the above startup of the target system, the execution monitor starts monitoring the arrival of every \emph{observation} (runtime event) as follows:
\begin{enumerate}
\item If the event is a ``method entry'', then the execution monitor checks if this is one of the ``entry events'' of the
corresponding component in the ``ready'' state; if so, then after receiving the data and the respective preconditions are checked; if they succeed, then the data is applied on the input port of the component and the mode of the execution state is changed to ``running''.
\item If the event is a ``method exit'', then the execution monitor checks if this one of the ``exit events'' of the component in the ``running'' state; if so, it changes its state into ``completed'' mode and collects the data from the output port of the component and checks for the corresponding postconditions. Should the checks fail, the execution monitor enables the diagnosis engine.
\item If the event is one of the ``allowable events'' of the component, it continues execution and finally
\item if the event is an unexpected event, i.e. it is neither an ``entry event'', nor an ``exit event'' and also not in ``allowable events'', then the execution monitor starts diagnosis.
\end{enumerate}

Based on the above behavioral description of the execution monitor, we have formalized the corresponding semantics of the execution monitor as follows:

\begin{tabbing}
$\forall$ \= app $\in$ Target\_System, sam $\in$ System\_Architectural\_Model, c $\in$ Component,
\\\> s, s' $\in$ State, t, t' $\in$ State$_s$, d, d' $\in$ Environment$_s$, e, e' $\in$ Environment, rte $\in$ RTEvent:
\\\> \textlbrackdbl sam\textrbrackdbl(d)(d', t, t') $\wedge$ \textlbrackdbl app\textrbrackdbl(e)(e', s, s') $\wedge$ startup(s, app) $\wedge$ isTop(c, \textlbrackdbl app\textrbrackdbl(e)(e', s, s')) $\wedge$
\\\> setMode(s, ``running'') $\wedge$ arrives(rte, s)  $\wedge$ equals(t, s) $\wedge$ equals(d, e)
\\\> $\Rightarrow$ \=
\\\>\> $\forall$ \= p, p' $\in$ Environment$^*$, m, n $\in$ State$_\bot^*$: 
\\\>\>\> equals(m(0), s) $\wedge$ equals(p(0), e)
\\\>\>\> $\Rightarrow$ \=
\\\>\>\>\> $\exists$ \= k $\in$ $\mathbb{N}$, p, p' $\in$ Environment$^*$, m, n $\in$ State$_\bot^*$:  
\\\>\>\>\>\> $\forall$ \=i $\in$ $\mathbb{N}_k$ : monitors(i, rte, c, p, p', m, n) $\wedge$
\\\>\>\>\>\>\> ( eq\=Mode(n(k), ``completed'') $\wedge$ eqFlag(n(k), ``normal'')  $\wedge$ 
\\\>\>\>\>\>\>\> equals(s', n(k))
\\\>\>\>\>\>\> $\vee$
\\\>\>\>\>\>\> eq\=Flag(n(k), ``compromised'') 
\\\>\>\>\>\>\>\> $\Rightarrow$ \=
\\\>\>\>\>\>\>\>\> enableDiagnosis(p'(k))(n(k), inBool(\texttt{true}))  $\wedge$ equals(s', n(k)) )
\end{tabbing}

The semantics of recursive monitoring is determined by two sequences
of states pre and post that are constructed from the
pre-state of the monitor. Any $ith$ recursion of the
monitor transforms $pre(i)$ state into $post(i+1)$ state from which the $pre(i+1)$
is constructed. No event can be accepted in an $Error$ state and the corresponding monitoring terminates either when the application has terminated with ``normal'' mode or when some misbehavior is detected as indicated by the respective ``compromised'' state. The corresponding ``monitors'' predicate
formalizes the aforementioned semantics as discussed in~\cite{MTK:2014}.
\enlargethispage*{4cm}

The formalization gives deep insight of the internal behavior of AWDRAT increasing its usability on the one hand and developing basis for its correctness (to be proved by automated tools) on the other hand.
Based on this formalism, we are currently working on the proof of the soundness of the Execution Monitor. The proof is essentially a structural induction proof, however, the non-trivial part is the recursive definition of the semantics of the monitor that is to be proved by the principle of rule induction~\cite{Winskel:1993}. We also plan to extend AWDRAT such that a target system's behavior is specified using Abstract State Machine~\cite{ASM:2003} based formalism which then will automatically translate into a semantically equivalent System Architectural Model allowing to already check the inconsistencies in the intra system behavior with various ASM automated tools, e.g. DKAL~\cite{DKAL:2013}.


%
\label{sect:bib}
\bibliography{sam}

\begin{thebibliography}{1}

\bibitem{ASM:2003}
E.~Borger and Robert~F. Stark.
\newblock {\em Abstract State Machines: A Method for High-Level System Design
  and Analysis}.
\newblock Springer-Verlag New York, Inc., Secaucus, NJ, USA, 2003.

\bibitem{DKAL:2013}
Jean-Baptiste Jeannin, Guido de~Caso, Juan Chen, Yuri Gurevich, Prasad Naldurg,
  and Nikhil Swamy.
\newblock {DKAL*: Constructing Executable Specifications of Authorization
  Protocols}.
\newblock Technical Report MSR-TR-2013-19, March 2013.

\bibitem{MTK12c}
Muhammad~Taimoor Khan and Wolfgang Schreiner.
\newblock {Towards the Formal Specification and Verification of Maple
  Programs}.
\newblock In Johan Jeuring, John~A. Campbell, Jacques Carette, Gabriel~Dos
  Reis, Petr Sojka, Makarius Wenzel, and Volker Sorge, editors, {\em
  Intelligent Computer Mathematics}, volume 7362 of {\em LNCS}, pages 231--247.
  Springer, 2012.

\bibitem{MTK:2014}
Muhammad~Taimoor Khan, Dimitrios Serpanos, and Howard Shrobe.
\newblock {On the Formal Semantics of the Cognitive Middleware AWDRAT}.
\newblock Technical Report CSAIL, MIT (to appear), September 2014.

\bibitem{Lamport:1994}
Leslie Lamport.
\newblock {The Temporal Logic of Actions}.
\newblock {\em ACM Trans. Program. Lang. Syst.}, 16(3):872--923, May 1994.

\bibitem{Shrobe:2006}
Howard Shrobe, Robert Laddaga, Bob Balzer, Neil Goldman, Dave Wile, Marcelo
  Tallis, Tim Hollebeek, and Alexander Egyed.
\newblock {AWDRAT: A Cognitive Middleware System for Information
  Survivability}.
\newblock In {\em Proceedings of the 18th Conference on Innovative Applications
  of Artificial Intelligence - Volume 2}, IAAI'06, pages 1836--1843. AAAI
  Press, 2006.

\bibitem{Shrobe:1979}
{Shrobe, Howard E}.
\newblock {Dependency Directed Reasoning for Complex Program Understanding}.
\newblock Technical report, Massachusetts Institute of Technology, Cambridge,
  MA, USA, 1979.

\bibitem{Winskel:1993}
Glynn Winskel.
\newblock {\em {The Formal Semantics of Programming Languages: An
  Introduction}}.
\newblock MIT Press, Cambridge, MA, USA, 1993.

\end{thebibliography}
\bibliographystyle{plain}



\end{document}